\documentclass{article} 
\usepackage{iclr2026_conference,times}

\usepackage{amsmath,amsfonts,bm}

\def\eqref#1{equation~\ref{#1}}

\def\1{\bm{1}}

\DeclareMathAlphabet{\mathsfit}{\encodingdefault}{\sfdefault}{m}{sl}
\SetMathAlphabet{\mathsfit}{bold}{\encodingdefault}{\sfdefault}{bx}{n}

\usepackage{hyperref}
\usepackage{url}
\usepackage{booktabs}
\usepackage{graphicx}
\usepackage{multirow}
\usepackage{multicol}
\usepackage{makecell}
\usepackage{pifont}
\usepackage{wrapfig}

\newcommand{\sysname}{STAR-VTON}
\newcommand{\varname}{VAR-VTON}
\title{Structure-Detail Decoupled Autoregressive Generation for Fast and High-Fidelity 
Virtual Try-On
}

\author{Lu Yang\textsuperscript{1}, Xiaonan Hu\textsuperscript{1}, Yanan Li\textsuperscript{2}, Daqi Liu\textsuperscript{3}, Hao Lu\textsuperscript{1,}\thanks{Corresponding author.}, Xiang Bai\textsuperscript{1}
\\
\textsuperscript{1}Huazhong University of Science and Technology, China\\
\textsuperscript{2}Wuhan Institute of Technology, China\\
\textsuperscript{3}Xiaomi EV, China\\
\texttt{\{lu\_yang1, hlu\}@hust.edu.cn} 
}

\iclrfinalcopy 
\begin{document}

\maketitle

\begin{abstract}

Virtual try-on (VTON) is a bi-conditional image generation problem that requires not only accurate person preservation but also faithful garment deformation and detail synthesis. 
Diffusion-based VTON methods can jointly model these factors in a compressed latent space, but suffer from high-frequency detail loss due to inherent latent compression, even with costly multi-step denoising.
Recent visual autoregressive (VAR) models offer a promising alternative for high-quality generation with faster inference, yet remain unexplored for VTON due to the lack of effective bi-conditioning mechanisms. 
To bridge this gap, we first introduce \varname{}, a VAR-based VTON model that incorporates garment conditioning and structural guidance for efficient latent-space VTON. 
Despite its efficacy, latent-space generation still struggles to preserve fine-grained garment details. We argue that different VTON sub-tasks should be addressed in different representation spaces: structural synthesis such as garment warping and person layout is suited to the latent space, whereas fine-grained detail recovery should be tackled in the pixel space.
Motivated by this insight, we further propose \sysname{}, a Two-Stage AutoRegressive framework that builds upon \varname{} by decoupling latent-space structural synthesis from pixel-space detail recovery. 
Our idea is to resort to a matching-informed refiner to establish dense correspondences between the stage-one generation and the source garment to directly map fine-grained pixel-space details. 
Extensive experiments show that \sysname{} achieves an impressive efficiency-fidelity trade-off: \varname{} runs at least $4\times$ faster than diffusion-based counterparts without degrading quality, and the pixel-space refiner effectively restores fine details and acts as a plug-and-play module that can benefit existing VTON approaches.

\end{abstract}

\section{Introduction}

\begin{wrapfigure}{r}{0.5\textwidth}
\centering
\vspace{-40pt}
\includegraphics[width=\linewidth]{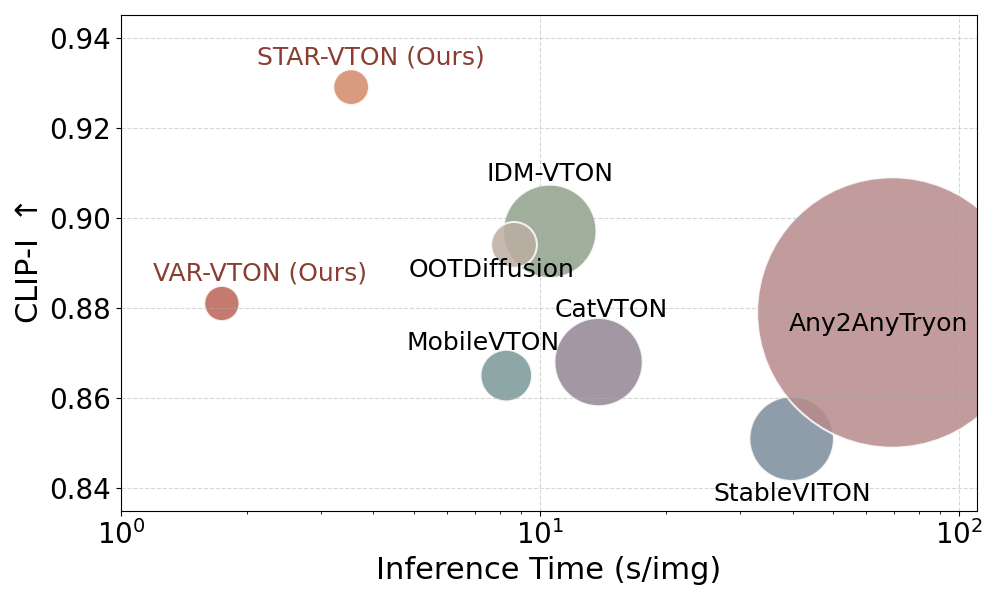}
\vspace{-20pt}
\caption{\textbf{Comparison with the state of the art in terms of detail fidelity and efficiency.} Bubble size denotes computational cost (TFLOPs). We evaluate detail preservation using CLIP-I,
with higher values indicating better fidelity. Our solutions achieve better fidelity-efficiency trade-off.
}
\vspace{-45pt}
\label{fig:abs}
\end{wrapfigure}

Given a person image and a garment image, virtual try-on (VTON) aims to synthesize a photorealistic image in which the person is naturally and accurately dressed in the target garment~\citep{han2018viton}. 
It is a typical bi-conditional generative problem, because a faithful VTON generation requires not only preserving the identity and pose of the person but also the fidelity of the garment.

Achieving high-fidelity VTON, however, is non-trivial. It needs to i) correctly deform the garment to align with the target body pose and shape, ii) well preserve person attributes including appearance and pose, and iii) faithfully retain garment details such as logos and text. 
Early GAN-based methods~\citep{han2018viton,wang2018toward,choi2021viton} typically follow a two-stage pipeline: a module first aligns garment deformation and human pose, followed by a generator~\citep{goodfellow2020generative} that synthesizes the try-on image. However, the reliance on explicit warping 
renders them vulnerable to large pose changes and 
non-rigid deformations.

\begin{figure}
    \centering
    \includegraphics[width=\linewidth]{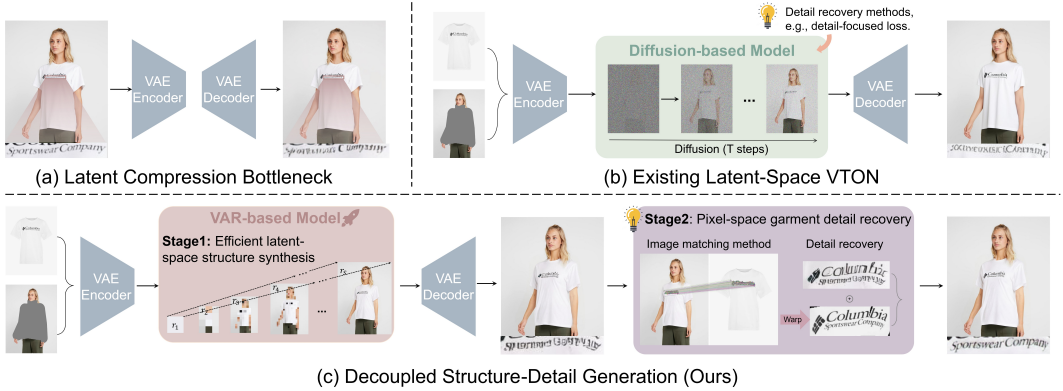}
    \vspace{-20pt}
    \caption{\textbf{Comparison between latent-space VTON and our structure-detail decoupled generation framework.}
(a) Existing VTON methods rely on compressed latent representations, where high-frequency garment details are inevitably lost during VAE encoding.
Consequently, (b) latent-space generation struggles to faithfully preserve fine-grained garment textures.
(c) Our framework leverages the VAR model for efficient holistic structure synthesis and dense image matching for pixel-space garment detail recovery, 
enabling both fast inference and faithful detail preservation.
}
    \label{fig:intro}
\end{figure}

Recent diffusion-based methods~\citep{kim2024stableviton,choi2024improving,chong2024catvton} alleviate the limitations above by jointly modeling garment deformation, person preservation, and garment-detail synthesis as a single generative process in compressed VAE latent space~\citep{kingma2013auto}, obviating explicit warping while yielding improved geometric alignment and photorealism.
Despite their visual appeal, these methods suffer from  i) high inference latency from multi-step denoising and ii) the inherent high-frequency detail loss due to VAE compression (Fig.~\ref{fig:intro}(a)).
Consequently, they struggle to maintain pixel-level detail fidelity, 
particularly for fine-grained patterns (Fig.~\ref{fig:intro}(b)).

Unlike diffusion models, visual autoregressive (VAR) models~\citep{tian2024visual,han2025infinity} have demonstrated high-quality image generation with substantially fewer decoding steps. 
Interestingly, this promising direction remains largely unexplored for efficient VTON. A possible reason is that VAR has not yet developed mature bi-conditional image generation toolkits such as ControlNet~\citep{zhang2023adding} and ReferenceNet~\citep{zhu2023tryondiffusion,hu2024animate}. To fill this gap, we first introduce \textbf{\varname{}}, a VAR-based VTON model for fast latent-space 
VTON, focusing on garment-person interaction and structural integrity. It is featured by i) a decoupled garment-prefix conditioning strategy that enables sufficient person-garment interaction during VAR generation, and ii) a scale-wise embedding mechanism of garment-agnostic representation that encourages the structural consistency of the target person.
However, we observe that our \varname{}, while more efficient, still fails to address the garment detail degradation (Fig.~\ref{fig:intro}(c)). 

The phenomenon above motivates us to rethink the current one-stage latent-space VTON. Indeed structural synthesis (e.g., pose alignment and deformation) is appropriately tackled in the compact latent space, but high-frequency details cannot be well recovered without accessing uncompressed pixel-level features. In this work, we return to the previous two-stage paradigm. Instead of 
burdening a single latent-space generator to resolve all challenges simultaneously, we 
posit that \textit{different sub-tasks should be tackled in representation spaces suited to their intrinsic properties}.
This motivates us to reformulate VTON as a structure-detail decoupled generation problem. To this end, we propose \textbf{\sysname{}}, a novel two-stage VTON approach based on  \varname{}, which decouples efficient structural synthesis from faithful garment-detail recovery, as illustrated in Fig.~\ref{fig:intro}(c).

Given the stage-one \sysname{} generation, we focus on pixel-space garment detail recovery in the second stage and introduce dense matching~\citep{shen2024gim} to recover high-frequency garment details from the original garment image. Specifically, we propose a matching-informed refiner that identifies texture-rich regions and 
dynamically injects warped garment details into the stage-one generation. By transferring garment details directly in pixel space, our method effectively preserves thin texts, small logos, and intricate patterns that are typically lost in latent-space generation.

Extensive experiments demonstrate that \sysname{} achieves a remarkable trade-off between generation fidelity and efficiency. Specifically, \varname{} runs over $4 \times$ faster than diffusion-based VTON counterparts while maintaining competitive generation quality, and the proposed pixel-space refiner consistently sets a new state of the art in garment-detail preservation on both VITON-HD~\citep{choi2021viton} and DressCode~\citep{morelli2022dress}. Notably, the refiner is model-agnostic and can be readily integrated into existing VTON methods, consistently improving detail preservation across different generative backbones. Overall, our results suggest that a two-stage framework that explicitly assigns structural synthesis and detail recovery to their respective representation spaces provides a new paradigm for designing efficient and high-fidelity virtual try-on.

\section{Related Work}
Our work is related to image-based VTON, controllable VAR generation, and dense image matching.

\subsection{Image-based Virtual Try-On}

Early image-based VTON approaches~\citep{han2018viton,wang2018toward,yu2019vtnfp,choi2021viton,lee2022high} commonly adopt a two-stage pipeline by first warping the garment to the target pose and then synthesizing the try-on image~\citep{goodfellow2020generative}. But inaccurate warping often leads to structural distortion under large pose variations.

Recently diffusion-based models~\citep{ho2020denoising,peebles2023scalable} have renewed the field with a unified generative process. 
Within this paradigm, a primary research focus has been dedicated to resolving the challenge of detail preservation. Early diffusion-based methods such as LaDI-VTON~\citep{morelli2023ladi} and DCI-VTON~\citep{gou2023taming} leverage CLIP~\citep{radford2021learning} embeddings to encode garment semantics, but these embeddings prioritize high-level concepts and fail to retain fine-grained textures. To mitigate this, StableVITON~\citep{kim2024stableviton} adopts a ControlNet-like~\cite{zhang2023adding} architecture, and dual-stream designs~\citep{zhu2023tryondiffusion,choi2024improving,xu2025ootdiffusion,zhou2025learning,wan2025incorporating} process the source garment and the target person independently to improve 
garment fidelity. However, parallel streams introduce substantial computational overhead; CatVTON~\citep{chong2024catvton} and TPD~\citep{yang2024texture} therefore consolidate feature extraction into a single-stream U-Net to improve efficiency. In parallel with detail enhancement, accelerating inference has also received much attention, e.g., MobileVTON~\citep{wan2026mobile} reduces computational cost through model distillation, and CAT-DM~\citep{zeng2024cat} addresses the speed bottleneck by initializing the reverse diffusion process with a GAN-generated implicit distribution, thereby drastically reducing the number of sampling steps without sacrificing visual quality.

Despite these attempts to jointly optimize detail retention and inference efficiency, existing pipelines remain hindered by their fundamental architectural constraints. In particular, the iterative denoising procedure imposes a strict lower bound on generation speed, while the reliance on the bottlenecked VAE latent space inevitably discards high-frequency information during compression. 
To overcome these limitations, we explore a decoupled generation framework. We leverage next-scale VAR prediction~\citep{tian2024visual} for rapid global structure synthesis, and establish 
direct correspondences in the uncompressed pixel domain to recover garment details.

\subsection{Controllable Visual Autoregressive Learning}

Recent 
VAR models~\citep{tian2024visual,han2025infinity} have achieved competitive image generation quality with 
far fewer decoding steps than diffusion models, making them attractive for efficient generation. To enable conditional control, existing methods typically inject guidance either at the pre-filling stage or during decoding.

Pre-filling based methods~\citep{li2024controlvar,qu2025varsr,dong2026echogen} prepend conditional tokens before 
VAR generation, enabling 
interaction between conditioning signals and generated tokens, which is suitable for appearance conditions that are spatially misaligned with the target image. In contrast, decoding-phase methods~\citep{li2024controlar,yao2024car,xu2026scalar} inject control signals throughout generation, providing stronger structural consistency when the conditioning signal is spatially aligned with the target image, such as pose maps or segmentation layouts.

VTON requires both strong garment-person interaction and faithful preservation of body structure. To satisfy both requirements, we adopt a hybrid conditioning strategy that injects garment appearance during pre-filling and garment-agnostic structural cues during decoding, enabling both human-garment interaction and person-consistent generation.

\subsection{Dense Image Matching} 

Recent advances in dense image matching have significantly improved the robustness of pixel-level correspondence 
under large viewpoint changes and non-rigid deformations~\citep{sun2021loftr,edstedt2023dkm,shen2024gim}. In particular, generalizable matching frameworks trained on large-scale video datasets~\citep{shen2024gim} have demonstrated a remarkable ability to establish reliable dense correspondences across diverse appearances and geometric transformations. These developments make it increasingly feasible to recover fine-grained visual details through correspondence-based transfer rather than generative synthesis.

Motivated by this opportunity, we revisit garment-detail preservation from a correspondence perspective. Rather than relying solely on generative models to synthesize fine-grained textures, we leverage dense image matching to directly recover garment details from the source image. This perspective bypasses the information bottleneck of latent-space generation and opens a new direction for preserving garment texture fidelity in VTON.

\section{\sysname{}: Two-Stage Autoregressive Virtual Try-On}
We begin with the problem formulation and then present the technical details of \sysname{}.

\subsection{Problem Formulation}

Given an image of a target person $\bm{x}_p$ and an image of a source garment $\bm{x}_g$, we extract a set of garment-agnostic representations $\mathcal{A}=\{\bm{m}, \bm{x}_m, \bm{x}_d\}$ from $\bm{x}_p$, where $\bm{m}$, $\bm{x}_m$, and $\bm{x}_d$ denote the garment mask, masked person image, and dense pose, respectively. In the first stage, 
structure synthesis is modeled via a VAR framework. A multi-scale vector-quantized (VQ)-encoder processes $\bm{x}_p$ to produce $S$ discrete token maps at 
progressively increased scales, denoted by $(\bm{r}_1,\dots,\bm{r}_S)$. 
Conditioned on the representation set $\mathcal{A}$ and the garment $\bm{x}_g$, a generator predicts 
the token maps in a coarse-to-fine manner. This process commences with the generation of the $1\times1$ map $\bm{r}_1$ and sequentially produces the map of each subsequent scale, that is,
\begin{equation}
p(\bm{r}_1,\dots,\bm{r}_S \mid \mathcal{A}, \bm{x}_g) = \prod_{s=1}^{S} p(\bm{r}_s \mid \bm{r}_1,\dots,\bm{r}_{s-1}, \mathcal{A}, \bm{x}_g) \,.
\end{equation}
A VQ-decoder then maps the generated sequence of tokens back to the pixel space, yielding an intermediate try-on result $\hat{\bm{x}}_{tr}$. While $\hat{\bm{x}}_{tr}$ generally captures the 
global topology and semantic coherence, it inherently lacks high-frequency 
details due to latent compression. Therefore, in the second stage, a pixel-level refinement process 
$f$ is introduced to 
recover these fine-grained details directly from the uncompressed garment $\bm{x}_g$, yielding the final high-fidelity try-on generation $\bm{x}_{tr}$ by
\begin{equation}
    \bm{x}_{tr} = f(\hat{\bm{x}}_{tr}, \bm{x}_g) \,.
\end{equation}

\begin{figure}[!t]
  \centering
  \includegraphics[width=\linewidth]{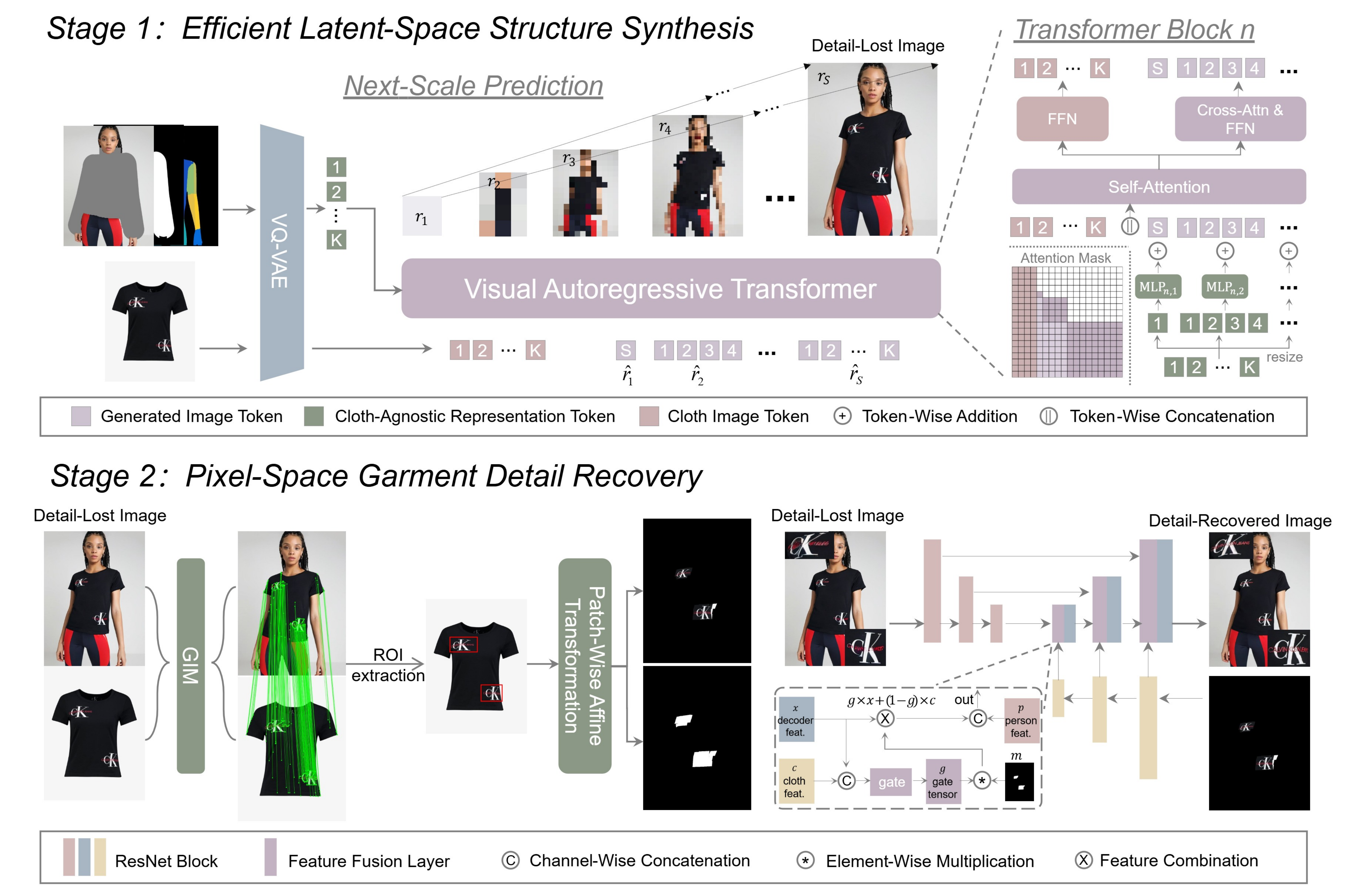}
    \caption{\textbf{Overview of \sysname{}.} Our method adopts a two-stage framework. 
    The first stage consists of two components: i) \textit{decoupled garment prefix conditioning}, which fuses garment tokens and image tokens within self-attention while maintaining separate processing pathways thereafter; and ii) \textit{scale-wise garment-agnostic representation embedding}, where multi-scale garment-agnostic features are progressively integrated into the generated tokens via dedicated MLPs at each scale and block. The second stage employs iii) a \textit{matching-informed refiner}, which establishes dense correspondences between the source garment and the generated output to recover details. }
  \label{fig:method}
\end{figure}
\subsection{Technical Challenges and Overview}
Applying VAR to virtual try-on is not a straightforward replacement of diffusion denoising with autoregressive token prediction. Unlike text- or class-conditioned generation, VTON requires jointly modeling two image-derived conditions with distinct roles: the source garment provides transferable appearance, while the garment-agnostic representation specifies the target body layout. This introduces two VAR-specific challenges: i) the garment should be globally accessible throughout autoregressive generation without being entangled with the target token sequence, and ii) structural errors introduced at coarse scales can propagate and accumulate during multi-scale decoding.

We propose \sysname{}, a two-stage structure-detail decoupled framework for efficient and high-fidelity virtual try-on.
To handle the VAR-specific issues discussed above, the first stage performs latent-space structure synthesis with two mechanisms:
i) \textit{decoupled garment prefix conditioning}, which uses garment tokens as condition prefixes to guide multi-scale token prediction; and
ii) \textit{scale-wise garment-agnostic representation embedding}, which injects target-person structural cues across autoregressive scales to maintain geometric consistency.
The second stage further overcomes the latent compression bottleneck in fine-grained detail preservation through iii) a \textit{matching-informed refiner}, which establishes dense correspondences between the original garment $\bm{x}_g$ and the intermediate VAR output $\hat{\bm{x}}_{tr}$, and selectively transfers high-frequency details to produce the final high-fidelity VTON result.
The overall pipeline is illustrated in Fig.~\ref{fig:method}.

\subsection{Latent-Space Autoregressive Structure Synthesis}

\paragraph{Decoupled Garment Prefix Conditioning.}
The garment information is injected in a prefix-based manner. Instead of directly concatenating the garment tokens and the multi-scale image tokens at the input end, 
we concatenate them only before the self-attention layer. After self-attention, the two sequences are separated: the image tokens proceed to the subsequent cross-attention and FFN, while the garment tokens are processed by 
another FFN. This decoupled encoding path ensures that garment information remains unaffected by text conditioning. Within self-attention, we design a causal mask (Fig.~\ref{fig:method}). It allows the generated tokens to attend unrestrictedly to the garment tokens, thereby extracting visual cues as needed. At the same time, it prevents garment tokens from attending to the generated tokens---a constraint essential for preserving the autoregressive sampling trajectory.

\vspace{-5pt}
\paragraph{Scale-Wise Garment-Agnostic Representation Embedding.}
SCALAR~\citep{xu2026scalar} and ControlAR~\citep{li2024controlar} show that for spatially aligned generation tasks, simple additive conditioning is sufficient for effective control. Following this observation, we inject garment-agnostic representations into the VAR block through scale-specific embeddings. Specifically, the masked person image $\bm{x}_m$ and dense pose $\bm{x}_d$ are encoded into latent variables $\bm{z}_m$ and $\bm{z}_d$ via a VQ-VAE, while the mask $\bm{m}$ is downsampled to match the resolution of the latent codes. The masked image latent $\bm{z}_m$, the dense pose latent $\bm{z}_d$, and the resized mask $\bm{m}$ are concatenated along the channel dimension to form the conditioning representation $\mathcal{A}$. To match the length of the generated tokens, $\mathcal{A}$ is resized to multiple scales. For each VAR block and at each scale of the generated tokens, a dedicated MLP processes the correspondingly scaled garment-agnostic representation, and the output is then added to the generated tokens. Formally,
\begin{equation}
\mathcal{A} = \bm{z}_m \textcircled{c} \bm{z}_d \textcircled{c} \bm{m}, \quad {\mathcal{A}_{r_1}, \mathcal{A}_{r_2}, \dots, \mathcal{A}_{r_S}} = \text{Resize}(\mathcal{A}) \,,
\end{equation}
\begin{equation}
\bm{x}_{i,j} = \bm{x}_{i,j} + \text{MLP}_{i,j}(\mathcal{A}_{r_j}) \,,
\end{equation}
where $\bm{x}_{i,j}$ denotes the token at the $j$-th scale output by the $i$-th block, $\mathcal{A}_{r_j}$ denotes the garment-agnostic representation at the $j$-th scale, and $\textcircled{c}$ denotes channel-wise concatenation.

\subsection{Pixel-Space Matching-Informed Detail Recovery}

We introduce a matching-informed refiner to recover lost details in VAE latents. The pixel correspondences between $\hat{\bm{x}}_{tr}$ and $\bm{x}_g$ are initially established using GIM~\citep{shen2024gim}. GIM produces a dense matching field $\Phi: \mathbb{R}^2 \to \mathbb{R}^2$, which maps garment pixels to their corresponding locations in the try-on image and thus provides the basis for pixel-space detail transfer.
Nevertheless, using $\Phi$ to warp the entire garment is prone to misalignment, especially in textureless, occluded, or heavily deformed regions.
We therefore use the matching points not only for warping but also for selecting reliable regions to warp.
Specifically, we construct a density map on the source garment from the matching points using Gaussian kernels and extract the high-density regions via thresholding, as the high-density regions generally indicate the areas that require textural refinement.
To achieve precise alignment with the try-on image, each selected region is further partitioned into multiple small patches that are warped independently and then composited. This process is formalized as
\begin{equation}
\bm{c}_w, \bm{c}_m = \sum_{p=1}^P{\text{Warp}(\bm{c}_p, \Phi)}\, ,
\end{equation}
where $P$ denotes the number of patches, $\bm{c}_p$ is the $p$-th garment patch, $\bm{c}_w$ is the warped garment, and $\bm{c}_m$ is the corresponding mask. 
An interesting phenomenon is that, even though textural details are lost in the try-on result, image matching remains accurate to produce precise garment warping.

Based on the warped garment regions, we employ a lightweight dual-encoder U-Net refiner to selectively inject the recovered pixel-space details into the stage-one generation. The refiner takes the intermediate try-on result $\hat{\bm{x}}_{tr}$ together with the warped garment $\bm{c}_w$ as input. Specifically, the person encoder extracts a set of multi-scale structural features
$\{\bm{f}_p^{l}\}_{l=1}^{L}$ from $\hat{\bm{x}}_{tr}$, while the garment encoder extracts
$\{\bm{f}_c^{l}\}_{l=1}^{L}$ from $\bm{c}_w$, where $L$ denotes the number of decoding stages.  The mask $\bm{c}_m$ is resized into a multi-scale pyramid $\{\bm{c}_m^l\}_{l=1}^{L}$ to constrain texture injection at each scale.
At the $l$-th decoding stage, let $\bm{f}_{dec}^l$ be the decoder feature. We update $\bm{f}_{dec}^l$ via a gating mechanism :
\begin{equation}
\bm{g}^{l}=\sigma\!\left(\text{Conv}\!\left(\text{Concat}[\bm{f}_c^{l},\bm{f}_{dec}^{l}]\right)\right) \in [0, 1]^{h \times w} \, ,
\end{equation}
\begin{equation}
    \bm{g}^{l}\leftarrow\bm{g}^{l}\odot\bm{c}_m^{l} \, ,
\end{equation}
\begin{equation}
    \tilde{\bm{f}}_{dec}^{l}=\bm{f}_c^{l}\odot\bm{g}^{l}+\bm{f}_{dec}^{l}\odot(1-\bm{g}^{l}) \, ,
\end{equation}

where $\sigma$ is the sigmoid activation and $\odot$ denotes element-wise multiplication.
The gate provides a spatially adaptive reliability measure for texture injection.
Instead of directly pasting the warped garment features, it allows the decoder to selectively borrow high-frequency details from $\bm{f}_c^l$ only where the warped texture is reliable, while preserving the synthesized structure in $\bm{f}_{dec}^l$ where local misalignment may occur.
The mask $\bm{c}_m^l$ further confines the gating operation to the matched garment regions, preventing irrelevant background or unreliable pixels from affecting the generated person structure.
The updated feature $\tilde{\bm{f}}_{dec}^{l}$ is then concatenated with the person skip feature $\bm{f}_p^l$ and propagated to the next decoding stage.
Training details are provided in the appendix.

\begin{table}[t]
    \centering
    \caption{Quantitative comparison on VITON-HD and DressCode. * denotes metrics computed on the detail-rich subset using garment-region crops, as described in Sec.~\ref{sec:experimental setup}. Inference latency is measured on an RTX 3090 GPU using bf16 precision. }
    \vspace{5pt}
    \addtolength{\tabcolsep}{-4pt}
    \renewcommand{\arraystretch}{1.1}
    \label{tab:vton_comprehensive}
    \resizebox{\textwidth}{!}{
        \begin{tabular}{lccc cc ccc cc ccc}
            \toprule
            \multirow{3}{*}{Method} & \multicolumn{5}{c}{VITON-HD} & \multicolumn{5}{c}{DressCode} & \multirow{3}{*}{\small Steps} & \multirow{3}{*}{\small TFLOPs} & \multirow{3}{*}{\makecell{\small Latency \\ (s/img)}} \\
            \cmidrule(lr){2-6} \cmidrule(lr){7-11}
            & \multicolumn{3}{c}{Paired} & \multicolumn{2}{c}{Unpaired} & \multicolumn{3}{c}{Paired} & \multicolumn{2}{c}{Unpaired} & & & \\
            \cmidrule(lr){2-4} \cmidrule(lr){5-6} \cmidrule(lr){7-9} \cmidrule(lr){10-11}
            & \small FID $\downarrow$ &\small KID $\downarrow$ &\small CLIP-I* $\uparrow$  &\small FID $\downarrow$ &\small KID $\downarrow$ &\small FID $\downarrow$ &\small KID $\downarrow$ &\small CLIP-I* $\uparrow$  &\small FID $\downarrow$\small & KID $\downarrow$ & & & \\
            \midrule
            \small StableVITON~\citep{kim2024stableviton}  & 6.39 & 1.11 & 0.851  & 9.30  & 1.42 & -    & -    & -       & -    & -    & 50 & 370.5 & 39.81 \\
            \small IDM-VTON~\citep{choi2024improving}     & 6.15 & 1.07 & \underline{0.897}  & 9.24  & 1.32 & 3.59 & 1.06 & 0.909  & \underline{5.46} & \underline{1.46} & 30 & 451.9 & 10.54 \\
            \small OOTDiffusion~\cite{xu2025ootdiffusion} & 6.22 & 1.79 & 0.894  & 9.25  & 1.99 & 4.03 & 0.83 & 0.909  & 7.56 & 2.59 & 20  & 108.7 &  8.66 \\
            \small CatVTON~\citep{chong2024catvton}      & 7.27 & 1.68 & 0.868  & 9.24  & 1.81 & 3.96 & 1.02 & 0.897  & \textbf{5.34} & \textbf{1.07} & 50 & 404.0 & 13.78  \\
            \small Any2AnyTryon~\citep{guo2025any2anytryon} & 8.08 & 1.74 & 0.879  & 10.17 & 2.20 & -    & -       & -    & 7.61 & 2.56 & 30 & 3791.1& 69.20 \\
            \small MobileVTON~\citep{wan2026mobile}   & 6.74 & 1.79 & 0.865  & 10.49 & 2.22 & -    & -    & -      & -    & -    & 28  & 138.1 & 8.30 \\
            \midrule
            \small \textbf{\varname{} (Ours)} & \underline{5.86} & \underline{0.85} & 0.881  & \underline{8.80} & \underline{0.86} & \underline{3.35} & \underline{0.74} & \underline{0.913}  & 5.62 & 1.61 & \textbf{13} & \textbf{64.4} &  \textbf{1.74}\\
            \small \textbf{\sysname{} (Ours)} & \textbf{5.84} & \textbf{0.84} & \textbf{0.929}  & \textbf{8.80} & \textbf{0.85} & \textbf{3.33} & \textbf{0.74} & \textbf{0.930}  & 5.56 & 1.57 & \underline{14} & \underline{66.3} &  \underline{3.54}\\
            \bottomrule
        \end{tabular}
    }
\end{table}
\section{Experiments}
Here we present our experimental setup, results, and discussion.
\subsection{Experimental Setup}
\label{sec:experimental setup}

\paragraph{Datasets.} We evaluate our method on VITON-HD~\citep{choi2021viton} and DressCode~\citep{morelli2022dress}.
VITON-HD contains $13,679$ upper-body image pairs, split into $11,647$ training pairs and $2,032$ testing pairs.
DressCode contains $53,795$ pairs from three categories: $15,366$ upper-body garments, $8,951$ lower-body garments, and $29,478$ dresses.
We follow the official split of DressCode, using $1,800$ pairs per category for testing and the remaining pairs for training.

\vspace{-10pt}
\paragraph{Implementation details.}
The VAR model initializes weights from the Infinity~\citep{han2025infinity}. Training proceeds in two stages: first at a resolution of $512\times 384$ for 10k steps, then at $1024\times 768$ for 15k steps. We train the refiner on $2,000$ detail-rich image pairs from VITON-HD. 
All training is conducted on 4 NVIDIA A6000 GPUs (48G memory each) with the AdamW optimizer and a batch size of 16. The learning rate is set to $1\times10^{-5}$. 

\vspace{-10pt}
\paragraph{Evaluation.}
We evaluate overall image quality in both paired and unpaired settings using Fréchet Inception Distance (FID)~\citep{heusel2017gans} and Kernel Inception Distance (KID)~\citep{binkowski2018demystifying}. To specifically assess garment-detail preservation, we construct detail-rich subsets from both VITON-HD and DressCode and measure the CLIP Image Similarity (CLIP-I)~\cite{hessel2021clipscore} between the generated and ground-truth garments. To focus exclusively on texture recovery, CLIP-I is computed only within the garment region. In addition, we report Structural Similarity Index (SSIM)~\citep{wang2004image} and Learned Perceptual Image Patch Similarity (LPIPS)~\citep{zhang2018unreasonable} on paired samples in the appendix as supplementary evaluations.

\vspace{-10pt}
\paragraph{Baselines.} We select StableVITON~\citep{kim2024stableviton}, IDM-VTON~\citep{choi2024improving}, OOTDiffusion~\citep{xu2025ootdiffusion}, CatVTON~\cite{chong2024catvton}, Any2AnyTryon~\citep{guo2025any2anytryon}, and MobileVTON~\citep{wan2026mobile} as baselines. For evaluation, each method generates try-on images at $1024\times768$ resolution when supported; otherwise, we generate results at $512\times384$, upscale them to $1024\times768$ using interpolation or super-resolution~\citep{wang2021real}, and report the best performance.
For fair latency comparison, all methods are evaluated with bf16 precision on a single RTX 3090 GPU, using the official sampling steps and averaging inference time over $10$ images.

\begin{figure}[!t]
    \centering
    \includegraphics[width=\textwidth]{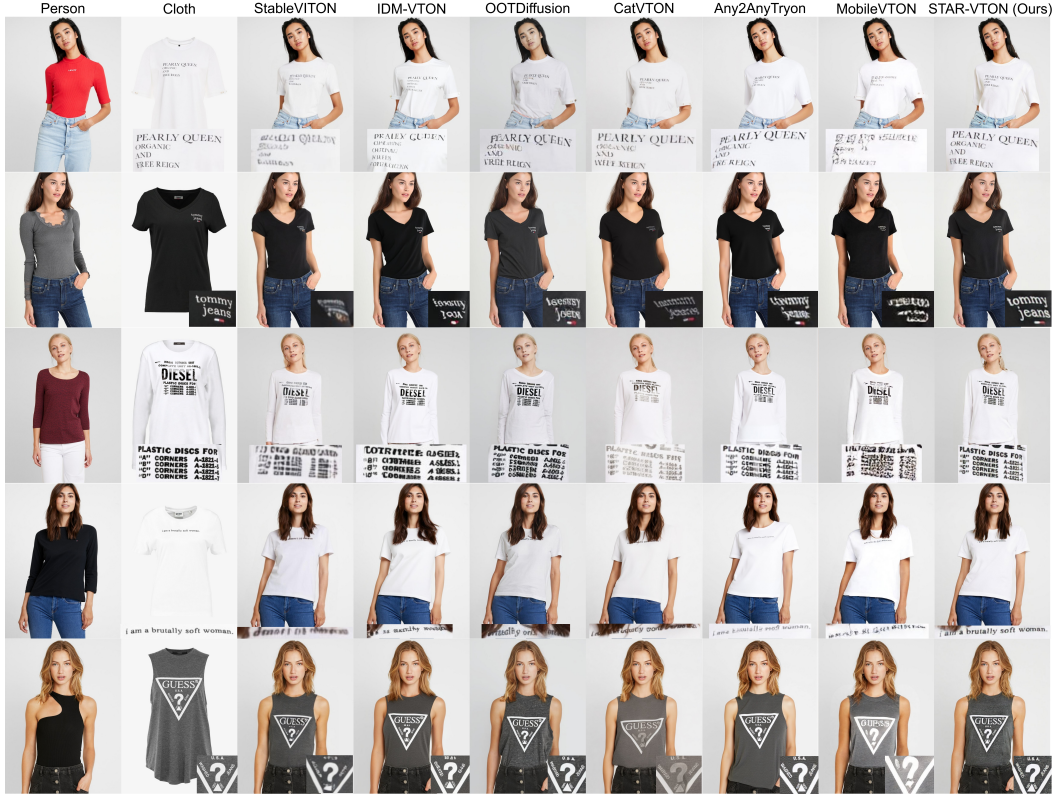}
    \caption{Qualitative comparison with baselines. Our method faithfully preserves fine-grained garment details, such as logos, text, and intricate patterns, that are often blurred or lost by existing methods. Zoom in for better viewing.}
    \label{fig:compare}
\end{figure}

\subsection{Main Results}

Table~\ref{tab:vton_comprehensive} and Fig.~\ref{fig:abs} present quantitative comparisons on VITON-HD and DressCode. Overall, \sysname{} achieves a superior trade-off between generation fidelity and efficiency. The results demonstrate that autoregressive generation serves as a strong alternative to diffusion models for 
VTON, achieving competitive or even superior image quality while significantly reducing computational cost and inference latency.
We additionally report results at $512\times384$ resolution in the appendix for a more comprehensive comparison.

Beyond efficiency, the structure-detail decoupled design further improves fine-grained garment preservation.
As shown in Table~\ref{tab:vton_comprehensive} and Fig.~\ref{fig:compare}, the pixel-space refiner enhances texture fidelity without degrading the generation quality of the latent-space synthesis stage.
The higher CLIP-I score of \sysname{} further demonstrates more faithful preservation of garment appearance, especially fine-grained textures and visual details.
These results support our hypothesis that global structure synthesis and detail recovery are better handled in latent and pixel spaces, respectively.


Furthermore, appendix experiments show that applying the proposed refiner to existing VTON methods consistently improves texture fidelity across existing VTON backbones, demonstrating the complementarity of pixel-space detail recovery and supporting our latent--pixel decoupling formulation.

\subsection{Ablation Study}
\label{sec:ablations}
\paragraph{Ablation on Garment Condition Embedding Strategy.}
We compare different garment condition embedding strategies in \varname{}.
As shown in Table~\ref{tab:cloth_injection}, the decoding-phase conditioning strategy of SCALAR~\citep{xu2026scalar} performs worse, indicating that additive injection is insufficient for spatially misaligned garment appearance.
Prefix conditioning improves all metrics by enabling direct interaction between garment tokens and generated tokens within self-attention.
Our decoupled prefix design achieves the best performance, since it protects garment representations from subsequent conditional interference and allows more accurate appearance modeling.

\vspace{-10pt}
\paragraph{Ablation on Garment-Agnostic Condition Embedding Strategy.}
We study different parameter-sharing strategies for garment-agnostic embedding in \varname{}.
As shown in Table~\ref{tab:person_injection}, using only block-specific or only scale-specific MLPs leads to inferior performance, while combining both performs best.
This indicates that structural cues play different roles across VAR scales and depths, making scale- and block-specific projections more effective for preserving person structure and pose.
\begin{table}[t]
  \centering
  \begin{minipage}[t]{0.49\linewidth}
    \centering
    \renewcommand{\arraystretch}{1.06}
    \caption{Ablation for garment condition embedding strategy on VITON-HD ($512\times384$).}
    \label{tab:cloth_injection}
    \resizebox{\textwidth}{!}{\begin{tabular}{lcccc}
      \toprule
      embedding Strategy & FID$\downarrow$ & KID$\downarrow$ & SSIM$\uparrow$ & LPIPS$\downarrow$ \\
      \midrule
      SCALAR                  & 6.59 & 0.42 & 0.854 & 0.070 \\
      prefix             & 5.87 & 0.35 & 0.852 & 0.065 \\
      decoupled prefix           & \textbf{5.22} & \textbf{0.28} & \textbf{0.864} & \textbf{0.054} \\
      \bottomrule
    \end{tabular}
    }
  \end{minipage}
  \hfill
  \begin{minipage}[t]{0.49\linewidth}
    \centering
    \addtolength{\tabcolsep}{-2pt}
    \caption{Ablation for garment-agnostic condition embedding strategy on VITON-HD ($512\times384$).}
    \label{tab:person_injection}
    \resizebox{\textwidth}{!}{\begin{tabular}{cc|cccc}
      \toprule
      Scale mlp & Block mlp & FID$\downarrow$ & KID$\downarrow$ & SSIM$\uparrow$ & LPIPS$\downarrow$ \\
      \midrule
      \ding{55} & \ding{51} & 5.57 & 0.41 & 0.861 & 0.057 \\
      \ding{51} & \ding{55} & 5.56 & 0.34 & 0.862 & 0.058 \\
      \ding{51} & \ding{51} &\textbf{5.22} & \textbf{0.28 }& \textbf{0.864} & \textbf{0.054}\\
      \bottomrule
    \end{tabular}
    }
  \end{minipage}
\end{table}

\begin{table}[t]
  \centering
  \begin{minipage}[t]{0.45\linewidth}
  \centering
  \caption{Ablation for garment encoding module on VITON-HD ($512\times384$).}
  \label{tab:cloth_encoder}
  \resizebox{\textwidth}{!}{\begin{tabular}{lcccc}
    \toprule
    Encoder & FID$\downarrow$ & KID$\downarrow$ & SSIM$\uparrow$ & LPIPS$\downarrow$ \\
    \midrule
    Trainable conv                & 10.21 & 1.57 & 0.845& 0.106 \\
    VQ-VAE                & \textbf{5.22} & \textbf{0.28} & \textbf{0.864} & \textbf{0.054} \\
    \bottomrule
  \end{tabular}
  }
  \end{minipage}
  \hfill
  \begin{minipage}[t]{0.54\linewidth}
  \centering
  \caption{Ablation for classifier-free guidance on VITON-HD ($1024\times768$).}
  \label{tab:ablation_cfg}
  \resizebox{\textwidth}{!}{\begin{tabular}{lccccc}
    \toprule
    CFG Strategy & FID$\downarrow$ & KID$\downarrow$ & SSIM$\uparrow$ & LPIPS$\downarrow$ & $t$ (s/img)$\downarrow$ \\
    \midrule
    w/ CFG                  & \textbf{5.75} & \textbf{0.80} & \textbf{0.878} & \textbf{0.073} & 2.48 \\
    \textbf{w/o CFG} & 5.86 & 0.85 & 0.876 & 0.076 & \textbf{1.74} \\
    \bottomrule
  \end{tabular}
  }
\end{minipage}
\end{table}

\begin{figure}[!t]
    \centering
    \includegraphics[width=\linewidth]{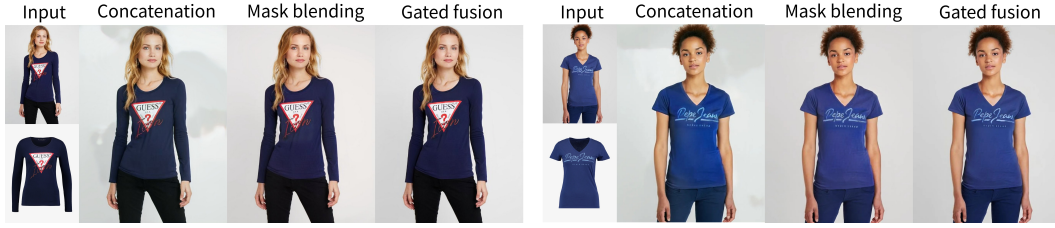}
    \caption{\textbf{Ablation on the gating strategy in the refiner.} Direct concatenation causes noticeable color shifts, while mask blending introduces obvious pasted artifacts. In contrast, the proposed gating mechanism produces more natural and coherent garment-detail recovery. Zoom in to compare.}
    \label{fig:gate_ablation}
\end{figure}

\vspace{-10pt}
\paragraph{Ablation on Garment Encoding Module.}
Table~\ref{tab:cloth_encoder} compares garment encoding strategies in \varname{}.
A trainable convolutional encoder performs notably worse than the pretrained VQ-VAE encoder, indicating that VQ-VAE latent features are better suited for autoregressive VTON generation.

\vspace{-10pt}
\paragraph{Ablation on Classifier-Free Guidance.}
Table~\ref{tab:ablation_cfg} evaluates the effect of classifier-free guidance (CFG). Although CFG brings moderate improvements in generation quality, it also increases inference latency. Since our primary goal is efficient virtual try-on generation, we adopt the setting without CFG as the default configuration.

\vspace{-10pt}
\paragraph{Ablation on Gating Strategy.}
We design three feature fusion strategies for the proposed refiner: i) \emph{Direct concatenation}, which directly concatenates the garment features $\bm{f}_c$ and decoder features $\bm{f}_{dec}$ along the channel dimension; ii) \emph{Mask blending}, which directly combines the two features using the warped garment mask: $ \bm{f}_{dec} \leftarrow \bm{f}_c \cdot \bm{c}_m + \bm{f}_{dec} \cdot (1 - \bm{c}_m)$
and iii) the proposed \emph{Gated fusion}, which adaptively fuses the two features through the learned spatial gate. The results in Fig.~\ref{fig:gate_ablation} demonstrate the importance of adaptive spatial fusion for natural garment-detail recovery.

\section{Conclusion}
In this work, we present \sysname{}, a novel virtual try-on framework that reformulates VTON as a decoupled structure-detail generation problem. Instead of relying on a unified latent-space generator to simultaneously handle garment deformation, person preservation, and fine-grained texture synthesis, our method separates efficient structure synthesis from faithful garment-detail recovery. Specifically, we explore Visual Autoregressive (VAR) modeling for fast latent-space try-on generation, and introduce a matching-informed pixel-space refiner to directly recover high-frequency garment textures through dense correspondence transfer.
Extensive experiments demonstrate that the proposed framework significantly reduces inference cost while preserving fine-grained garment details with high fidelity. 
We hope this work highlights the importance of modeling different synthesis objectives in representation spaces that best match their intrinsic properties, and provides a new perspective for future research on efficient and high-fidelity virtual try-on generation.


\section*{Acknowledgments}
This work is jointly supported by the National Natural Science Foundation of China under Grant No. 62576146 and  No. 62441615.

\bibliography{iclr2026_conference}
\bibliographystyle{iclr2026_conference}

\newpage
\appendix
\section{Appendix}
\subsection{Training Details}
We train the VAR model following the configuration of the official Infinity~\citep{han2025infinity} repository. For the refiner, we select $2{,}000$ detail-rich image pairs from VITON-HD. The refiner is optimized using MSE loss $\mathcal{L}_{mse}$, perceptual loss~\citep{johnson2016perceptual} $\mathcal{L}_{p}$, adversarial loss $\mathcal{L}_{adv}$, and Discrete Fourier Transform (DFT) loss $\mathcal{L}_{fft}$. The overall objective is formulated as:
\begin{equation}
    \mathcal{L}=\lambda_{mse}\mathcal{L}_{mse}+\lambda_{p}\mathcal{L}_{p}+\lambda_{adv}\mathcal{L}_{adv}+\lambda_{fft}\mathcal{L}_{fft}\,,
\end{equation}
where $\lambda_{mse}$, $\lambda_{p}$, and $\lambda_{adv}$ are set to $1$, and $\lambda_{fft}$ is set to $0.01$.

To simulate the information loss introduced by latent compression, we first compress and reconstruct the images using VQ-VAE at a resolution of $512\times384$, and then upsample them to $1024\times768$ for refiner training. During inference, imperfect detail generation may cause mismatches between the warped garment patterns and the synthesized images. To improve robustness against such inconsistencies, we first train the refiner on aligned data and subsequently fine-tune it using training samples with slight distortions applied to the warped garment patterns.

\subsection{More Experimental Results}

\begin{wraptable}{r}{0.5\textwidth}
\centering
\vspace{-20pt}
\caption{Supplementary evaluation on paired VTON settings using structural similarity (SSIM) and perceptual similarity (LPIPS).  These metrics serve as complementary evaluation to the main results reported in Table~\ref{tab:vton_comprehensive}.}
\resizebox{\linewidth}{!}{
        \begin{tabular}{lcccc}
            \toprule
            \multirow{2}{*}{Method} & \multicolumn{2}{c}{VITON-HD} & \multicolumn{2}{c}{DressCode}  \\
            \cmidrule(lr){2-3} \cmidrule(lr){4-5}
            & \small SSIM $\uparrow$ &\small LPIPS $\downarrow$ &\small SSIM $\uparrow$ &\small LPIPS $\downarrow$  \\
            \midrule
            \small StableVITON~\citep{kim2024stableviton}  &\underline{0.890} & 0.077 & - &-  \\
            \small IDM-VTON~\citep{choi2024improving}     & 0.869 &\underline{0.075} & 0.902 &\textbf{0.052} \\
            \small OOTDiffusion~\cite{xu2025ootdiffusion} &0.839 & 0.087 & 0.884 & 0.069  \\
            \small CatVTON~\citep{chong2024catvton}      & 0.875 & 0.096 & 0.899 & 0.073  \\
            \small Any2AnyTryon~\citep{guo2025any2anytryon} & 0.870 & 0.112 & - & -  \\
            \small MobileVTON~\citep{wan2026mobile}   & \textbf{0.896 }& \textbf{0.068} &- &- \\
            \midrule
            \small \textbf{VAR-VTON (Ours) }& 0.876 & 0.076 &\underline{0.903} &\underline{0.056} \\
            \small \textbf{STAR-VTON (Ours)} & 0.876 & 0.076 &\textbf{0.903} & 0.057 \\
            
            \bottomrule
        \end{tabular}
    }
    \label{tab:ssim}
    \vspace{-10pt}
\end{wraptable}

\paragraph{Additional paired evaluation with SSIM and LPIPS.} To further complement the main evaluation, we additionally report Structural Similarity Index (SSIM)~\citep{wang2004image} and Learned Perceptual Image Patch Similarity (LPIPS)~\citep{zhang2018unreasonable} on paired VTON settings, as shown in Table~\ref{tab:ssim}. These metrics are commonly used to assess structural consistency and perceptual similarity between generated results and ground-truth images. In our work, they serve as supplementary indicators to the primary FID/KID and CLIP-I evaluations, providing an additional perspective on image-level reconstruction quality.

\vspace{-10pt}
\paragraph{Comparison at $512\times384$ resolution.}
For fair comparison with prior methods evaluated at lower resolutions, we additionally report results on VITON-HD at $512\times384$ resolution in Table~\ref{tab:vton_comparison}. \varname{} is trained and evaluated directly at $512\times384$, while for baselines supporting $1024\times768$ generation, we first generate high-resolution results and then downsample them to $512\times384$. Even under this setting, \varname{} achieves the best overall performance while maintaining substantially lower computational cost and faster inference speed. These results further demonstrate the effectiveness and efficiency of the proposed VAR-based global synthesis framework.

\begin{figure}[!t]
    \centering
    \includegraphics[width=\linewidth]{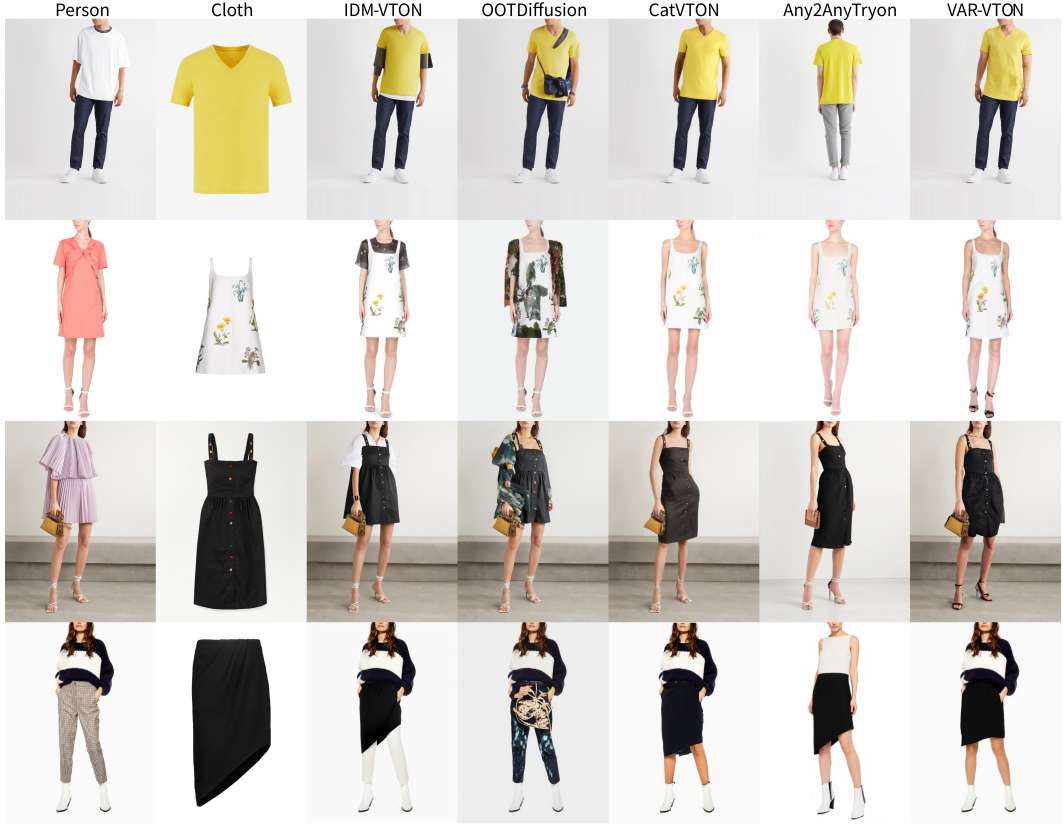}
    \caption{\textbf{Qualitative comparison on the DressCode dataset.} Our method produces realistic garment deformation across diverse garment categories.}
    \label{fig:dresscode_compare}
\end{figure}

\vspace{-10pt}
\paragraph{Additional comparison on DressCode.}
We further provide additional qualitative comparisons on the DressCode dataset in Fig.~\ref{fig:dresscode_compare}. These results suggest that the proposed VAR-based global synthesis framework generalizes effectively beyond VITON-HD and can achieve competitive synthesis quality even for more challenging garments and poses.

\begin{figure}[!t]
    \centering
    \includegraphics[width=\linewidth]{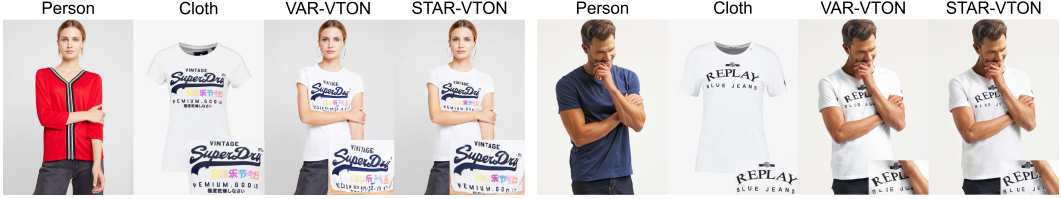}
    \caption{\textbf{Refiner results under challenging poses with partial occlusion.}}
    \label{fig:complex_pose}
\end{figure}

\vspace{-10pt}
\paragraph{Refiner performance under challenging poses.} We further evaluate the robustness of the proposed pixel-space refiner under challenging poses and partial occlusions. As shown in Fig.~\ref{fig:complex_pose}, when garment regions are partially occluded or distorted due to complex human poses, the refiner is still able to recover high-frequency textures from visible regions by leveraging dense image correspondences between the synthesized result and the source garment.
This demonstrates that the refiner does not simply perform local sharpening, but instead relies on matching-informed texture transfer, enabling adaptive restoration of unoccluded garment details even under significant geometric deformation. These results further highlight the generalization capability of the proposed pixel-space refinement mechanism beyond standard frontal or well-aligned cases.

\begin{figure}[!t]
    \centering
    \includegraphics[width=\linewidth]{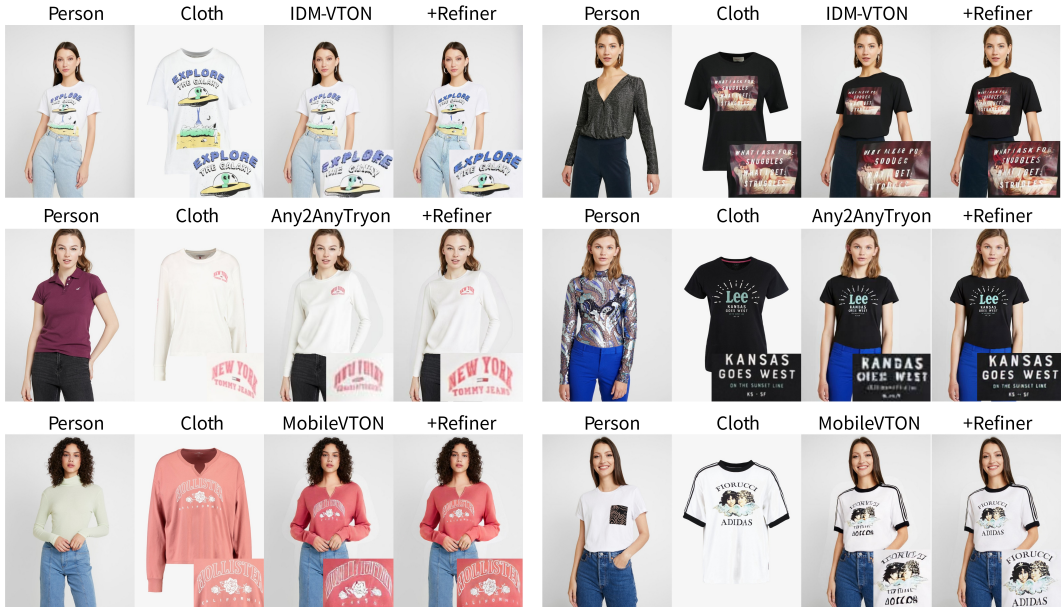}
    \caption{\textbf{Compatibility of the proposed refiner with different baseline methods.} The proposed refiner consistently improves the preservation of fine-grained garment textures, while maintaining the overall appearance generated by the original models.}
    \label{fig:refiner_w_other}
\end{figure}

\vspace{-10pt}
\paragraph{Compatibility of the refiner.}
To further validate the generality of the proposed refiner, we apply it to the outputs of several existing VTON methods, as shown in Fig.~\ref{fig:refiner_w_other}. The refiner consistently enhances fine-grained garment details, including small logos and textual patterns, while preserving the global structure generated by the original models.
We further provide quantitative results in Table~\ref{tab:refiner}, where the proposed refiner is applied to multiple baselines across VITON-HD and DressCode. As shown, the refinement module consistently improves CLIP-I scores across all methods and datasets, demonstrating that the proposed pixel-space refinement is model-agnostic and can serve as a general plug-and-play enhancement for improving garment-detail fidelity.

\begin{table}[t]
    \centering
    \caption{Quantitative comparison on VITON-HD ($512 \times 384$ resolution)}
    \addtolength{\tabcolsep}{-4pt}
    \renewcommand{\arraystretch}{1.1}
    \label{tab:vton_comparison}
        \begin{tabular}{lcccc cc ccc}
            \toprule
            \multirow{2}{*}{Method} & \multicolumn{4}{c}{Paired} & \multicolumn{2}{c}{Unpaired} & \multirow{2}{*}{\small Steps}&\multirow{2}{*}{\small TFLOPs} &\multirow{2}{*}{\makecell{\small Latency \\ (s/img)}} \\
            \cmidrule(lr){2-5} \cmidrule(lr){6-7}
            & \small FID $\downarrow$ &\small KID $\downarrow$ &\small SSIM $\uparrow$ &\small LPIPS $\downarrow$ &\small FID $\downarrow$ &\small KID $\downarrow$  & & \\
            \midrule
            \small StableVITON~\citep{kim2024stableviton}  & 6.44 & 0.94 & 0.854 & 0.091 & 11.05  & 3.91  & 50 & 92.8 & 8.91 \\
            \small IDM-VTON~\citep{choi2024improving}     & \underline{5.76} & \underline{0.73} & 0.850 & 0.060 & 9.84  & 1.12  & 30 & 117.4 & 5.85 \\
            \small OOTDiffusion~\cite{xu2025ootdiffusion} & 9.31 & 4.09 & 0.819 & 0.088 & 12.41  & 4.69  & \underline{20}  & \underline{27.3} &  \underline{1.14} \\
            \small CatVTON~\citep{chong2024catvton}      & 5.43 & 0.41 & 0.870 & 0.057 & 9.02  & 1.09  & 50 & 101.0 & 2.59  \\
            \small Any2AnyTryon~\citep{guo2025any2anytryon} & 6.93 & 0.76 & 0.838 & 0.088 & \underline{8.79} & \underline{0.89}  & 30 & 1091.4& 17.91 \\
            \small MobileVTON~\citep{wan2026mobile}   & 5.95 & 1.07 & \textbf{0.892} & \textbf{0.049} & 9.75 & 1.66     & 28  & - & - \\
            \midrule
            \small \textbf{\varname{} (Ours)} & \textbf{5.22} & \textbf{0.28} & \underline{0.864} & \underline{0.054} & \textbf{8.34} & \textbf{0.47} & \textbf{10} & \textbf{19.9} &  \textbf{0.78}\\
            \bottomrule
        \end{tabular}
\end{table}

\begin{table}[!t]
    \centering
    \caption{Effectiveness of the proposed pixel-space refiner on existing VTON methods. We apply the refiner to multiple baseline models and report \textbf{CLIP-I} scores.}
    \addtolength{\tabcolsep}{-4pt}
    \renewcommand{\arraystretch}{1.1}
    \label{tab:refiner}
    \resizebox{\textwidth}{!}{
    \begin{tabular}{lccccccc}
    \toprule
     \multicolumn{2}{l}{Datasets \& Methods} & StableVITON & IDM-VTON & OOTDiffusion & CatVTON & Any2AnyTryon & MobileVTON \\
     \midrule
     \multirow{2}{*}{VITON-HD}    &Latent-space synthesis & 0.851 & 0.897 & 0.894 & 0.868 & 0.879 & 0.865 \\
    &  +Pixel-space refinement &0.923 & 0.920 &0.922 & 0.922 & 0.916 & 0.919 \\
     \midrule
     \multirow{2}{*}{DressCode}    &Latent-space synthesis & - & 0.918 & 0.909 & 0.897 & -& - \\
    & +Pixel-space refinement &- & 0.940 &0.926 & 0.921 & -& - \\
    \bottomrule

    \end{tabular}
}
\end{table}

\subsection{Limitations}

Although the proposed \sysname{} achieves strong efficiency and garment-detail preservation, several limitations remain. In this work we mainly focus on recovering fine-grained textures in pixel space, while the global synthesis stage primarily relies on a standard VAR formulation without dedicated detail-preservation mechanisms inside latent-space generation. As a result, although the proposed VAR-based framework achieves competitive overall synthesis quality, its intrinsic ability to preserve fine-grained textures within latent space is still weaker than diffusion-based models. Importantly, the proposed pixel-space refiner should not be viewed as a replacement for detail preservation during structure synthesis. The effectiveness of the refiner still depends on the quality of the generated try-on image. When the global synthesis stage fails to preserve sufficient garment details, the number and reliability of image matching correspondences decrease accordingly, which can negatively affect the subsequent pixel-space refinement process. Specifically, the refinement process becomes more sensitive to hyperparameters such as the number of warped patches. In some challenging cases, a single refinement stage may also be insufficient for fully recovering complex textures, potentially requiring iterative or multi-stage refinement strategies. We believe that jointly improving latent-space detail preservation and pixel-space refinement is an important direction for future work.

\end{document}